\documentclass{article}
\usepackage{ijcai25}      

\usepackage{times}        
\usepackage{soul}
\usepackage{url}
\usepackage[hidelinks]{hyperref}
\usepackage[utf8]{inputenc}
\usepackage[small]{caption}
\usepackage{graphicx}
\usepackage{amsmath}
\usepackage{amssymb}
\usepackage{booktabs}
\usepackage{algorithm}
\usepackage{algorithmic}
\usepackage[switch]{lineno} 
\usepackage{natbib} 
\usepackage{subfigure}

\title{A Survey on Data-Centric AI: Tabular Learning from Reinforcement Learning and Generative AI Perspective}

\author{
Wangyang Ying$^1$ \and
Cong Wei$^1$ \and
Nanxu Gong$^1$ \and
Xinyuan Wang$^1$ \and
Haoyue Bai$^1$ \and
Arun Vignesh Malarkkan$^1$ \and
Sixun Dong$^1$ \and
Dongjie Wang$^2$ \and
Denghui Zhang$^3$ \and
Yanjie Fu$^1$
\affiliations
$^1$Arizona State University \\
$^2$The University of Kansas \\
$^3$Stevens Institute of Technology
\emails
\{wying4, cwei35, nanxugong,xwang735, haoyueba, arun.malarkkan, sixundong, yanjie.fu\}@asu.edu,
wangdongjie@ku.edu, dzhang42@stevens.edu
}

\begin{document}
\maketitle

\begin{abstract}
Tabular data is one of the most widely used data formats across various domains such as bioinformatics, healthcare, and marketing. As artificial intelligence moves towards a data-centric perspective, improving data quality is essential for enhancing model performance in tabular data-driven applications.
This survey focuses on data-driven tabular data optimization, specifically exploring reinforcement learning (RL) and generative approaches for feature selection and feature generation as fundamental techniques for refining data spaces. Feature selection aims to identify and retain the most informative attributes, while feature generation constructs new features to better capture complex data patterns.
We systematically review existing generative methods for tabular data engineering, analyzing their latest advancements, real-world applications, and respective strengths and limitations. This survey emphasizes how RL-based and generative techniques contribute to the automation and intelligence of feature engineering.
Finally, we summarize the existing challenges and discuss future research directions, aiming to provide insights that drive continued innovation in this field.
\end{abstract}

\vspace{-0.4cm}
\section{Introduction}

In modern machine learning, models have become increasingly large and effective, but the high cost of GPUs presents a challenge. Data-centric AI (DCAI) offers an alternative by enhancing data quality and representation to improve AI performance, even when using simpler models~\cite{kumar2024opportunities}.
Among different data types, tabular data remains one of the most fundamental and widely used formats, with applications spanning healthcare, finance, marketing, etc~\cite{liu2024calorie,safriandono2024analyzing,lu2020traumatic}. Unlike unstructured data, such as images and text, tabular data is characterized by complex feature dependencies, high dimensionality, and stringent interpretability requirements. Additionally, its availability is often more restricted compared to vision and language datasets, making it challenging to develop generalizable AI solutions. As a key area of DCAI, tabular learning aims to transform a given tabular dataset into an optimal representation by leveraging feature knowledge convergence. These factors necessitate advanced transformation techniques that maximize the utility of tabular data.


Feature engineering is essential for extracting meaningful patterns from tabular datasets, significantly impacting the performance of machine learning models~\cite{heaton2016empirical}. It comprises two fundamental tasks: feature selection and feature generation. Feature selection~\cite{mrmr,biesiada2008feature,ding2014identification,lasso1996} aims to identify the most relevant features by eliminating redundancy and irrelevant information, thereby enhancing model interpretability, efficiency, and generalization. Feature generation~\cite{chen2021techniques,kusiak2001feature,kanter2015deep,azim2024feature}, on the other hand, focuses on constructing new features that capture complex interactions and domain-specific patterns, enriching the input representation for better predictive performance. 

Among these innovations, reinforcement learning (RL)~\cite{sutton2018reinforcement} and generative models~\cite{ruthotto2021introduction} have emerged as powerful frameworks for optimizing tabular learning. RL-based techniques employ reward-driven exploration to iteratively refine feature selection and transformation strategies, allowing models to learn adaptive feature representations that improve downstream predictive tasks. By dynamically exploring various feature combinations, RL enables data-driven optimization of feature spaces without the need for manual intervention. Generative models, in contrast, learn latent feature representations that facilitate intelligent feature construction, capturing intricate relationships within data. These models can generate synthetic features that preserve underlying statistical properties, thereby enhancing model robustness and generalization. Moreover, generative approaches enable systematic knowledge transfer by reusing learned feature representations across different datasets, reducing the need for extensive labeled data in new tasks.
\begin{figure*}[t]
    \centering
\includegraphics[width=1.0\linewidth]{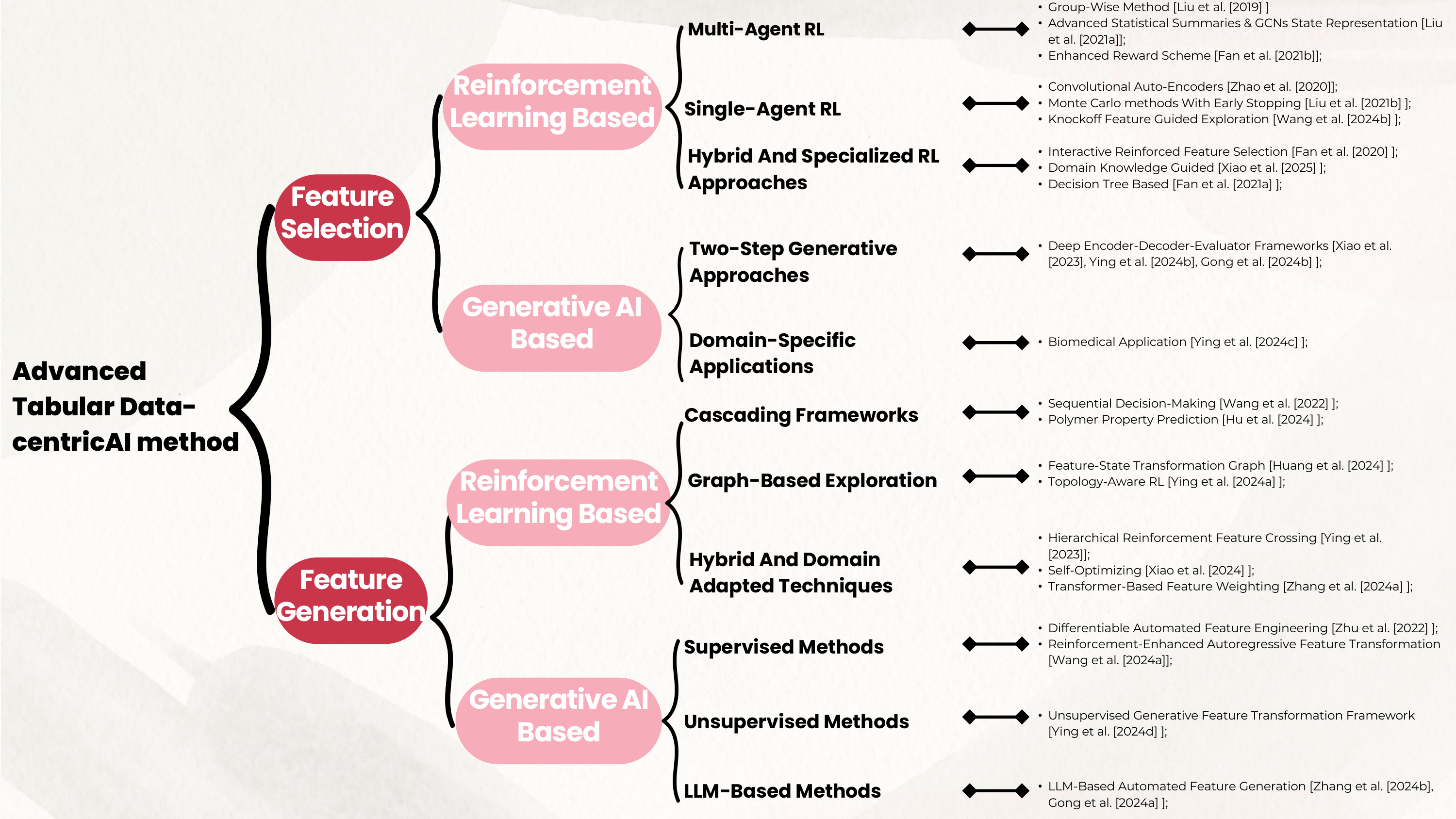}
    \vspace{-0.5cm}
    \caption{A taxonomy overview of RL-based and generative techniques in tabular data-centric AI.}
    \label{fig:overview}
    \vspace{-0.5cm}
\end{figure*}

To advance research in this area, this survey systematically reviews existing feature selection and generation techniques, with a particular focus on RL-based optimization and generative AI approaches (see \textbf{Figure~\ref{fig:overview}}). We explore the unique challenges posed by tabular data, examine how advanced AI methodologies address these challenges, and provide comparative insights into different transformation strategies. Additionally, we discuss emerging trends that are shaping the future of automated feature engineering. Through this comprehensive analysis, we aim to provide researchers and practitioners with a deeper understanding of cutting-edge methodologies and their implications for real-world applications.
\vspace{-0.2cm}
\section{Data-Centric AI: Tabular Learning}

Tabular data is one of the most common data formats in the real world, widely used in fields such as bioinformatics, healthcare, and marketing. It has a well-defined structure, where rows represent samples and columns represent features, allowing for precise representation of numerical and categorical information and their relationships. Additionally, tabular data is highly interpretable. However, its practical application faces several challenges, including many redundant features, complex feature interactions, the difficulty of automated modeling, and privacy concerns due to data sensitivity. These issues limit model efficiency and performance, hindering the widespread adoption of AI for tabular data.

To address these challenges, Data-Centric AI focuses on improving data quality and representation to enhance model performance, with feature selection and feature generation as its core tasks. Feature selection eliminates redundant and irrelevant information, retaining only the most valuable features for modeling, while feature generation constructs a more discriminative feature space, enabling models to learn complex patterns more effectively. By optimizing these aspects, Data-Centric AI reconstructs a refined and information-rich data representation, improving the practicality and reliability of tabular data in AI applications.

\textbf{Feature selection}  aims to identify and retain the most informative features while discarding redundant or irrelevant ones, thereby improving model performance, efficiency, and interoperability.

\textbf{Feature generation} aims to rebuild a new feature space from an original feature set (e.g. $[f_1, f_2] \rightarrow [\frac{f_1}{f_2}, f_1-f_2, \frac{f_1+f_2}{f_1}]$), where $f$ represents a feature (a.k.a, a column) of a tabular dataset. It can advance the power (structural, interaction, and expression levels) of data to make data AI-ready.

Since tabular data lacks inherent spatial or sequential structure, feature extraction is typically explicit, relying heavily on manual feature engineering techniques such as interaction term construction and aggregation. In contrast, unstructured data (e.g., images and audio) can leverage deep learning to automatically extract implicit features, whereas tabular data still faces challenges in automating feature construction.
Reinforcement learning (RL) and generative AI introduce new possibilities for automating and optimizing feature engineering in tabular data. RL enables intelligent search and optimization mechanisms for automated feature selection and transformation, reducing reliance on manual engineering. Meanwhile, generative AI can create new features or enhance existing ones, improving the expressiveness of tabular data.
Tabular data-centric AI aims to transform raw datasets into optimized representations that enhance interpretability, computational efficiency, and predictive accuracy. This approach bridges the gap between traditional manual feature engineering and automated techniques, offering a more efficient and intelligent solution for tabular data analysis.

\section{Reinforcement Learning Perspective for Data-Centric AI}

\begin{figure}[t]
  \centering
  \subfigure[\small{Feature Selection}]{
    \includegraphics[width=0.227\textwidth]{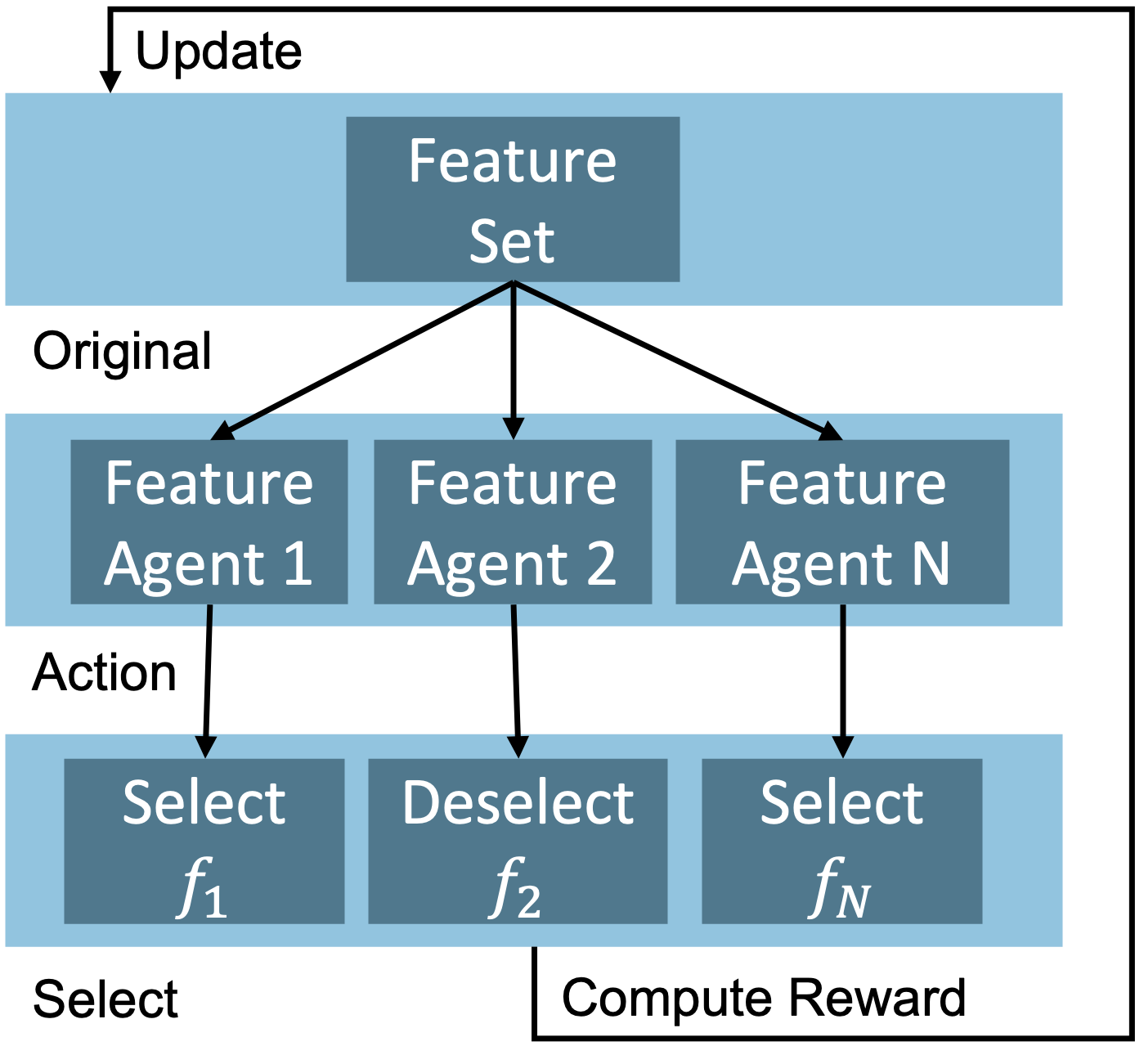}
  }
  \hfill
  \subfigure[\small{Feature Generation}]{
    \includegraphics[width=0.227\textwidth]{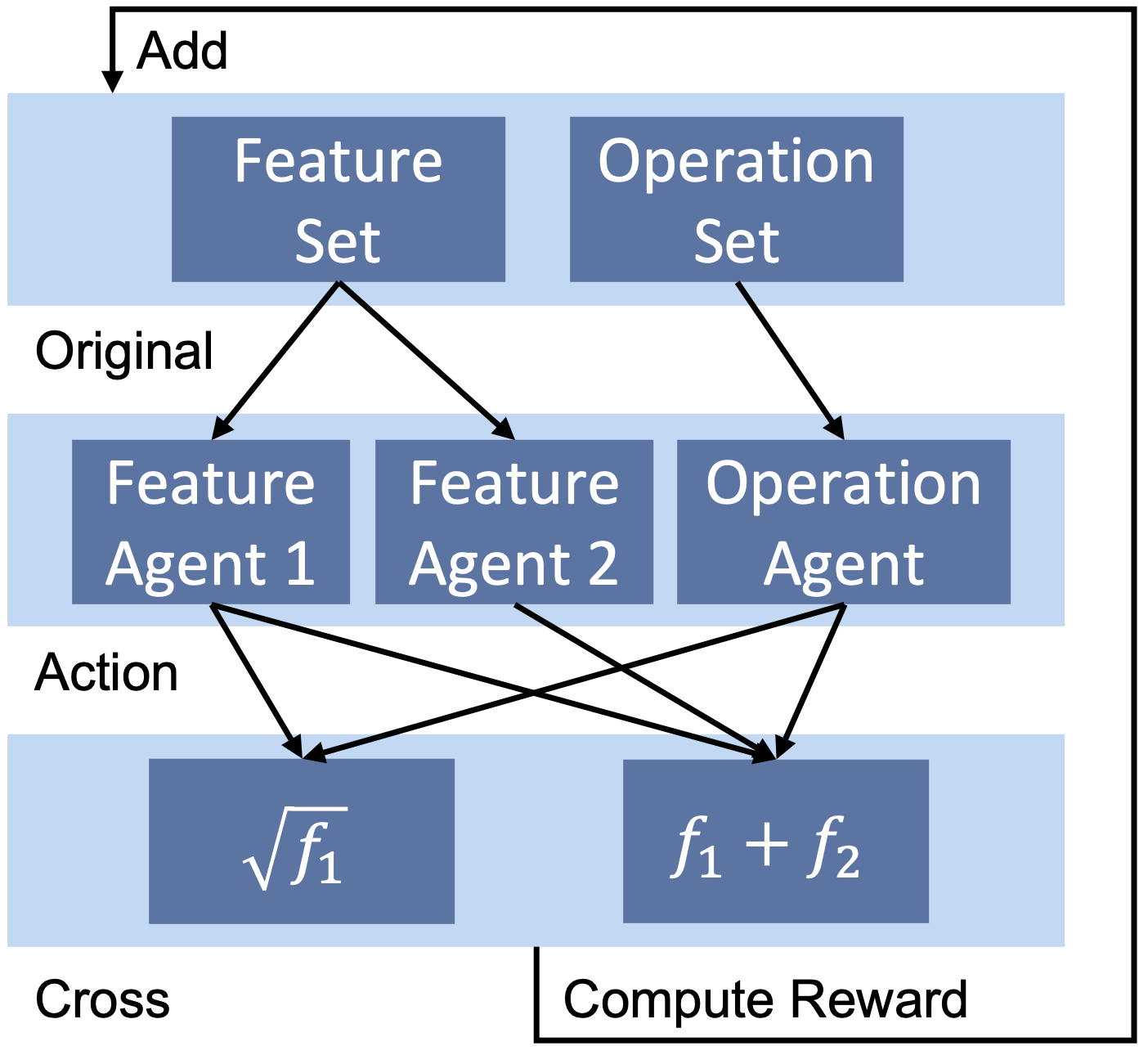}
  }
  \vspace{-0.3cm}
  \caption{Feature selection and feature generation are formulated as RL problems.}
  \vspace{-0.3cm}
  \label{fig:feat}
\end{figure}

Reinforcement learning (RL) guides feature selection and generation through a reward mechanism. Its exploratory search strategy closely aligns with the iterative process used by human experts to refine features, enabling it to effectively discover the most discriminative feature combinations (see \textbf{Figure~\ref{fig:feat}}).

\subsection{RL for Feature Selection}
RL formulates feature selection as a sequential decision process, where an agent selects informative features to maximize a reward based on model performance and feature relevance. Existing works can be categorized into three-fold: \textbf{Multi-Agent RL Frameworks}, \textbf{Single-Agent RL Frameworks}, and \textbf{Hybrid and Specialized RL Approaches}. 

The multi-agent framework enables more efficient large-scale feature selection tasks through parallel exploration. \cite{liu2019automating} proposed a framework where each feature is assigned an independent agent, leveraging advanced state representations, such as statistical summaries and Graph Convolutional Networks (GCNs), to improve feature selection intelligence. Furthermore, \cite{liu2021automated} optimized the reward mechanism to improve collaboration among agents. To further reduce computational resource requirements, \cite{fan2021autogfs} cluster similar features, and each group is managed by a single agent. This approach maintains scalability while significantly reducing computational overhead.

However, the multi-agent framework faces the challenge of high computational costs. To address this issue, \cite{zhao2020simplifying} proposed a single-agent approach, integrating all decisions into a single agent to enable sequential feature exploration, thereby reducing computational overhead. \cite{liu2021efficient} further improved efficiency by combining the single-agent framework with Monte Carlo methods and early stopping. Additionally, \cite{wang2024knockoff} introduced knockoff features to guide exploration, achieving more robust and effective feature selection.

Hybrid and specialized RL methods integrate RL with other techniques to adapt it to specific domains. \cite{fan2020autofs} incorporated an external trainer within the RL framework to guide RL agents, enhancing the breadth of exploration. \cite{xiao2025} leveraged biological domain knowledge to guide biomarker identification. \cite{fan2021interactive} utilized decision tree feedback to model the hierarchical structure of features, improving state representation and personalizing feature selection.

In conclusion, RL-driven feature selection provides a novel approach to tackling the challenges of high-dimensional and complex feature spaces. Through multi-agent collaboration, single-agent optimization, and hybrid integrations, these methods showcase RL's adaptability in refining feature selection processes, enhancing efficiency, and ultimately boosting machine learning model performance.

\subsection{RL for Feature Generation}
RL-based feature generation models feature construction as a sequential decision process, where an agent selects existing features, applies mathematical transformations, and optimizes the generated features for downstream tasks. Existing works can be categorized into \textbf{cascading frameworks}, \textbf{graph-based exploration}, and \textbf{hybrid and domain-adapted techniques}.
The cascaded RL framework formulates the feature generation process as a sequence of dependent decision steps, where each step builds upon the previous one.
\cite{wang2022group} introduced a three-agent cascaded framework, where two agents select features and one agent chooses an operator, iteratively constructing new features. Expanding on this idea, \cite{hu2024reinforcement} applied cascaded RL to polymer property prediction, using this structured approach to generate meaningful descriptors.

Graph-based methods leverage feature relationships to optimize feature generation. \cite{huang2024enhancing} constructs a feature state transition graph to track valuable transformations, enhancing the RL exploration strategy. \cite{ying2024topology} extracts core data and encodes features using graph neural networks, ensuring that latent structural information within the data is captured, thereby improving feature transformation capabilities.

Hybrid methods integrate RL with other relevant techniques to address domain-specific challenges. \cite{ying2023self} proposed a hierarchical reinforcement feature interaction approach, incorporating feature discretization, hashing, and descriptive summaries to enhance feature generation quality. \cite{xiao2022self} employed a self-optimizing framework to mitigate the issue of overestimated Q-values. \cite{zhang2024tfwt} leveraged a Transformer-based attention to capture feature dependencies, reducing data redundancy.

In conclusion, RL-based feature generation introduces a dynamic and adaptive approach to constructing features. By integrating cascaded decision-making, structural learning, and hybrid methodologies, these frameworks showcase RL's ability to automate and optimize feature engineering, paving the way for more efficient and insightful data representations.

\section{Generative Perspective for Data-Centric AI}
Generative AI, with its ability to encapsulate knowledge within a latent space, offers a systematic approach to compress feature information and generate enhanced feature representations. It can learn hidden patterns from high-dimensional, sparse, or complex non-linear tabular data, producing more expressive features than traditional methods.

\begin{figure}[t]
    \centering
\includegraphics[width=1.0\linewidth]{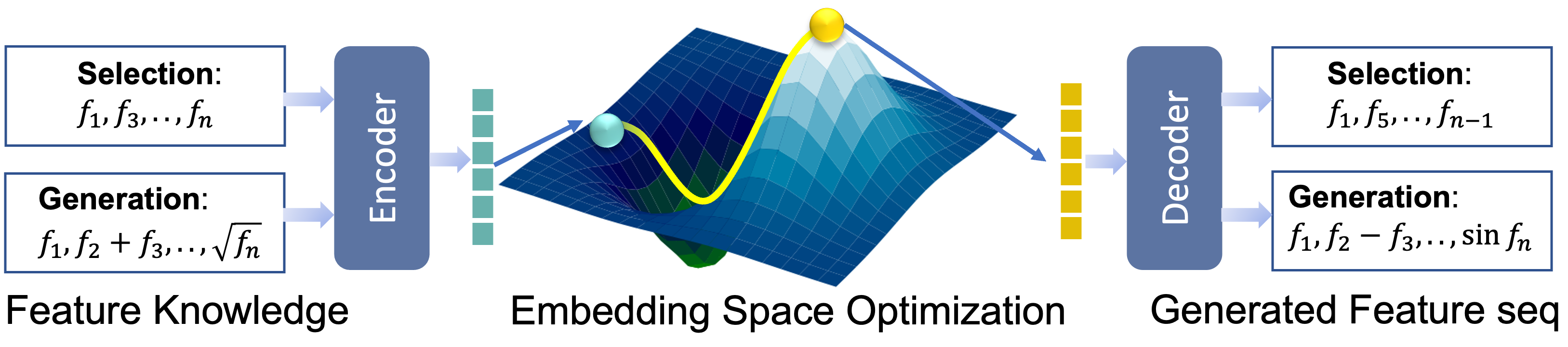}
    \vspace{-0.3cm}
    \caption{Generative AI captures feature knowledge in a continuous embedding space, enabling gradient-based search to identify optimal feature sets.}
    \label{fig:generative AI}
    \vspace{-0.3cm}
\end{figure}

\subsection{Generative AI for Feature Selection} 
Generative AI redefines feature selection by transforming it from a discrete search problem into an optimization problem in continuous space. As shown in \textbf{Figure~\ref{fig:generative AI}}, it embeds empirical feature selection behaviors as knowledge into a continuous space, and then generates selection decisions based on this space, thereby constructing a more effective feature selection framework.

~\cite{GAINS} introduces an encoder-decoder-evaluator framework that leverages reinforcement learning to automatically collect observed feature selection experiences, then uses an encoder to embed them into a continuous space, and finally generates new feature selection results by a decoder. This framework ensures the diversity of observed data and enhances the effectiveness of feature selection tasks using generative models.

~\cite{VTFS} redefines feature selection as a sequential token generation task based on the encoder-decoder-evaluator paradigm. This framework incorporates a transformer-based variational autoencoder (VAE), effectively capturing complex dependencies among features while mitigating model overfitting and sensitivity to noise.

Building upon this, ~\cite{FNFS} further investigates feature redundancy in high-dimensional feature spaces. This framework evaluates feature orthogonality and integrates it into the generative framework, identifying a more compact but more precise feature subset while minimizing redundancy in the selection results.

Moreover, generative AI-based feature selection has been applied to specific domains. ~\cite{GERBIL} identifies key biomarkers in high-dimensional, low-sample-size biological gene datasets, enhancing the effectiveness of early disease detection through genomic analysis. This highlights the transformative potential of generative AI in healthcare, where precise feature selection can directly improve disease prediction and reduce healthcare costs.

In conclusion, generative AI methods significantly enhance the effectiveness, stability, and generalization capability of feature selection while demonstrating substantial potential across various application domains. These advancements pave the way for broader research and applications in high-dimensional data environments.

\subsection{Generative AI for Feature Generation}
Generative AI constructs an optimal feature space through the embedding-optimization-generation paradigm. The core idea is to embed generated feature sets as observed feature knowledge into a continuous space to learn feature interactions, optimize within this space, and ultimately generate an optimal set of features. This paradigm has demonstrated significant potential in feature generation tasks, effectively improving model performance while reducing manual intervention.

~\cite{zhu2022difer} randomly generates generated feature sets as observed feature knowledge to guide feature generation. However, the poor performance of randomly generated observations leads to ineffective embeddings, while the independent handling of embedding and generation processes introduces noise and errors. Additionally, manually setting the number of generated features and using a greedy search strategy results in suboptimal generation outcomes.

Building on this, ~\cite{NIPS@MOAT} incorporates RL into the framework. RL’s exploration-exploitation mechanism is used to automatically generate high-quality observation data, which is encoded using postfix expressions to optimize the construction of the embedding space. Within this space, gradients are used to locate the optimal embeddings, enabling the generation of superior feature generations. However, this framework heavily relies on feedback from downstream predictive tasks to search for optimal results, making it time-consuming and limiting its applicability in unsupervised learning scenarios.

To address this limitation, ~\cite{KDD@NEAT} proposes an unsupervised feature generation framework. This approach represents each observed generated feature set as a feature-feature similarity graph and employs graph contrastive learning and multi-objective fine-tuning to optimize the embedding space, thereby decoupling feature generation from downstream predictive tasks. However, the effectiveness of this framework is constrained by graph augmentation strategies and manually designed unsupervised utility evaluation metrics.

Recently, advanced methods leveraging large language models (LLMs) have expanded the possibilities of feature generation. ~\cite{zhang2024dynamic} introduces a dynamic and adaptive automatic feature generation process using LLMs. This method reconstructs the feature space based on feedback from downstream tasks, utilizing LLMs' context learning and reasoning capabilities to dynamically adjust and optimize the generated features across various machine learning scenarios.

Furthermore, ~\cite{gong2024evolutionary} proposes an evolutionary LLM framework for automated feature generation. This method formulates feature generation as a sequence generation task, integrating an RL-based data collector to construct a multi-population database. The LLM leverages few-shot learning to generate improved feature generation sequences, while evolutionary algorithms (EAs) further refine the search path, enhancing both the quality and generalizability of the feature generations.

In conclusion, generative AI has achieved significant advancements in feature generation, substantially improving the efficiency and quality of feature modeling. These methods address the inefficiencies of discrete search spaces and enhance feature interaction modeling, enabling AI to exhibit greater adaptability. Future research directions could explore smarter adaptive optimization strategies, integrate multi-modal data (e.g., text, images, time series) for feature generation, and further enhance the automation and generalization capabilities of feature engineering.

\section{Strengths and Limitations of RL-based vs. Generative Methods}

RL-based feature engineering leverages reinforcement learning to automatically optimize feature space, improving data quality and model adaptability, while generative AI synthesizes high-quality data or augments key features to mitigate data scarcity. Both approaches enhance the core capabilities of data-centric AI, making data quality optimization a key driver of AI performance. We summarize the key differences between these two approaches in terms of performance, interpretability, adaptability, and automation.

\noindent
\textbf{Performance.}
\begin{itemize}
    \item \emph{RL-based Methods:} RL-based methods optimize feature transformations by aligning them with the final task objective, making them effective for dynamically changing data. They excel in high-dimensional datasets by capturing nonlinear relationships and adapting features automatically. However, RL has high computational costs, slow convergence, and relies on well-designed reward functions. Poor reward design can lead to suboptimal solutions, limiting effectiveness.
    \item \emph{Generative-based Methods:} Generative-based methods reformulate feature engineering as a continuous optimization problem, allowing smoother transformations and more efficient search in high-dimensional spaces. This makes them more stable than RL in some cases but also dependent on high-quality training data. However, generative models may introduce biases, and their training remains computationally expensive, especially for complex datasets.
\end{itemize}

\noindent
\textbf{Interpretability.}
\begin{itemize}
    \item \emph{RL-based Methods:} RL-based methods offer better interpretability because they rely on discrete decision paths, making feature transformations traceable through policy networks or Q-values. However, as feature space complexity increases, deep RL models become harder to interpret, and their effectiveness depends on task-specific reward functions, which may not always provide clear reasoning for feature selection.
    \item \emph{Generative-based Methods:} Generative-based methods optimize features in a continuous space, allowing smoother transformations and potential insights from latent variable representations. However, their interpretability is limited due to the black-box nature of deep generative models, making it difficult to directly correlate generated features with input data patterns. High-dimensional probability modeling further complicates feature importance analysis.
\end{itemize}

\noindent
\textbf{Adaptability and Automation.}
\begin{itemize}
    \item \emph{RL-based Methods:} RL-based methods exhibit strong adaptability by dynamically adjusting feature selection and transformation strategies to match different data distributions. They are well-suited for rapidly changing environments, such as time-series or streaming data. However, their automation is constrained by the need for well-designed reward functions, as poor reward design can lead to suboptimal strategies. Additionally, RL relies on discrete search, which, while capable of exploring complex feature combinations, is often slow and less efficient for large-scale feature engineering.
    \item \emph{Generative-based Methods:} Generative-based methods optimize features in a continuous space, making them highly adaptable to various data patterns, especially in unsupervised or semi-supervised scenarios. Their automation level is higher than RL since generative models can directly learn latent data structures and synthesize high-quality features without predefined search strategies. However, their adaptability depends on training data quality—if data distributions shift significantly, the generated features may become invalid or biased. Moreover, training generative models can be complex, often requiring careful tuning of architectures and hyperparameters to perform well across different tasks.
\end{itemize}
\vspace{-0.4cm}
\section{Practical Strategies for Effective Feature Engineering}
Feature engineering plays a critical role in optimizing machine learning performance, especially when employing RL-based or generative-based approaches. While these methods offer automation and adaptability, their effectiveness depends on proper implementation. This section provides key strategies for designing robust feature engineering pipelines, balancing model complexity, domain expertise, explainability, scalability, and ethical considerations.

\noindent
\textbf{1) Start Simple Before Adopting Advanced Models.}
A common mistake in feature engineering is jumping directly to complex RL policies or high-capacity generative models. Instead, it is more effective to begin with straightforward methods and gradually increase complexity:
\begin{itemize}
    \item \emph{RL Starter Policy:} An initial policy can rely on simple reward signals, such as validation accuracy, before incorporating more sophisticated constraints like feature cost or domain-specific penalties. This stepwise refinement ensures that early-stage learning remains stable while progressively aligning with real-world objectives.
    \item \emph{Generative Bootstrapping:} it is beneficial to first apply basic transformations, such as linear autoencoders, to understand fundamental data patterns. More advanced architectures like GANs or Transformer-based generative models should only be introduced once the baseline has demonstrated meaningful feature improvements.
\end{itemize}

\noindent
\textbf{2) Incorporate Domain Knowledge for Smarter Decision-Making.}
While automation reduces manual effort, human expertise remains crucial for guiding feature engineering:
\begin{itemize}
    \item In RL, domain experts can shape the agent’s reward function and define an action space that prioritizes meaningful features. This prevents RL from selecting redundant or irrelevant transformations, improving both efficiency and interpretability.
    \item For generative-based approaches, experts can constrain latent variables or enforce certain transformation rules, ensuring that generated features align with practical needs. This is particularly important in fields like healthcare and finance, where feature validity must comply with industry regulations.
\end{itemize}

\noindent
\textbf{3) Ensure Continuous Validation and Interpretability.}
Feature engineering should be an iterative process, with ongoing evaluation of its impact on downstream tasks:
\begin{itemize}
    \item Newly derived features should be tested for their contribution to model accuracy, robustness, and generalization. Monitoring feature utilization can help identify redundant or misleading transformations.
    \item Interpretability tools such as local surrogate models (e.g., LIME~\cite{KDD@LIME}) or attribution techniques help explain how RL-based policies select features or how generative models manipulate latent spaces. This is particularly crucial in high-risk applications where transparency is required.
\end{itemize}

\noindent
\textbf{4) Address Scalability and Computational Constraints.}
Large-scale datasets introduce computational challenges, making it necessary to optimize resource allocation:
\begin{itemize}
    \item For RL-based methods, high-dimensional action spaces can lead to slow exploration and increased training costs. Techniques like hierarchical RL or offline RL can mitigate this issue by reducing search complexity and improving efficiency.
    \item For generative models, training on vast datasets with high-dimensional features can be computationally expensive. Instead of processing entire datasets, a more practical approach is to train on smaller subsets or compressed representations before scaling to full-sized data.
\end{itemize}

\noindent
\textbf{5) Maintain Ethical Standards and Regulatory Compliance.}
Automated feature engineering must align with privacy and ethical guidelines when dealing with sensitive data:
\begin{itemize}
    \item Privacy-preserving strategies should be applied when using generative models for sensitive datasets. Techniques such as differential privacy~\cite{baiprivacy} or federated learning~\cite{kamatchi2025securing} can prevent the unintentional leakage of confidential information.
    \item Regulatory compliance requires transparency—RL-based feature selection may produce transient or dynamic features that are difficult to track. Maintaining clear logs of selected transformations is essential for auditing, particularly in regulated industries.
\end{itemize}
\vspace{-0.2cm}
\section{Choosing RL-Based vs. Generative AI}
Selecting the right feature engineering approach depends on various factors, including the nature of the data, computational constraints, and the specific optimization objectives. RL-based methods and generative models offer distinct advantages, making them suitable for different scenarios. This section outlines key considerations for choosing between these two approaches and explores how they can be effectively combined.

\noindent
\textbf{RL-based Feature Engineering for Dynamic and Sequential Optimization.}
Reinforcement learning is particularly well-suited for scenarios where feature importance evolves over time or where decisions must be made sequentially. When working with streaming data or environments that require continuous adaptation, RL agents can iteratively refine feature transformations based on real-time feedback. This makes RL effective in applications such as online learning systems, reinforcement-based recommendation models, and adaptive data preprocessing pipelines. Additionally, RL-based methods excel in optimization tasks where incremental improvements are necessary, ensuring that feature selection aligns with long-term performance gains.

\noindent
\textbf{Generative AI for Large-Scale Feature Exploration and Static Datasets.}
Generative models are more appropriate when dealing with high-dimensional datasets where exploring numerous feature transformations simultaneously is beneficial. Unlike RL, which searches in a discrete space, generative models optimize features in a continuous space, allowing for smoother transformations and better scalability. These methods are particularly useful in batch-processing environments, where they can generate diverse feature representations that are later filtered and refined. Use cases include data augmentation, synthetic feature generation, and unsupervised discovery of latent patterns, making them effective for improving model performance when labeled data is limited.

\noindent
\textbf{Hybrid Approaches: Leveraging the Strengths of Both Methods.}
In some cases, combining RL-based and generative approaches can lead to superior performance. One strategy involves using generative models to create a large pool of candidate features, which an RL agent then evaluates and selects dynamically. Conversely, RL can optimize feature selection in streaming applications, while a generative model periodically retrains in an offline setting to propose new feature transformations. This hybrid strategy balances adaptability with computational efficiency, ensuring both short-term flexibility and long-term feature diversity.

\noindent
\textbf{Key Considerations for Choosing the Right Approach.}
The decision to use RL-based or generative feature engineering should be guided by several practical factors:
\begin{itemize}
    \item \emph{Nature of the Data:} If data is continuously changing, RL is the better choice. If the dataset is static or semi-static, generative methods may be more effective.
    \item \emph{Feature Space Complexity:} RL is useful when feature selection requires careful, iterative refinement. Generative AI is more efficient for searching through a vast feature space.
    \item \emph{Computational Resources:} RL training can be expensive due to the need for extensive exploration. Generative models, while also computationally demanding, can often be trained in a more controlled offline manner.
    \item \emph{Interpretability Requirements:} RL’s decision-making process is more transparent since it follows a structured selection path, whereas generative models may introduce complex transformations that are harder to explain.
\end{itemize}

\section{Challenges and Future Research Directions}

As tabular data-centric AI continues to evolve, researchers face numerous challenges while also encountering new opportunities for innovation. This chapter explores six key issues: automation in feature engineering, interpretability, privacy preservation, computational efficiency, scalability, and the integration of large language models (LLMs) and multimodal learning. For each aspect, we analyze the existing challenges and outline potential research directions to advance the field.

\noindent\textbf{Automation in Feature Engineering: Balancing Adaptability and Efficiency.}
Automating feature engineering has significantly improved data processing efficiency by reducing manual intervention. However, real-world applications often demand domain-specific customizations that automated methods struggle to accommodate. Furthermore, current approaches face scalability limitations when dealing with high-dimensional, large-scale datasets, leading to excessive computational costs. Future Research Directions:
1) Human-in-the-loop hybrid systems: Combining automated techniques with expert knowledge can create adaptable systems that maintain efficiency while incorporating domain-specific insights.
2) Resource-efficient strategies: Optimizing computational frameworks for automated feature engineering will enable their deployment in environments with limited processing power, increasing accessibility.

\noindent\textbf{Enhancing Interpretability: Making Feature Transformations Transparent.}
Feature engineering plays a crucial role in model performance, yet its processes are often opaque. In high-stakes domains such as healthcare, and finance~\cite{liu2025calorie,leng2023nlrp3}, a lack of transparency can undermine trust, making it essential to understand the rationale behind feature transformations. Future Research Directions:
1) Traceable transformation workflows: Developing transparent feature engineering pipelines will allow users to track each transformation step and its relationship with model predictions.
2) Quantifying feature impact: Establishing methods to measure the influence of feature transformations on performance helps identify the most critical steps, fostering confidence in AI-driven decisions.

\noindent\textbf{Privacy-Preserving Feature Engineering.}
With growing concerns about data privacy, federated learning has emerged as a promising approach for distributed feature engineering. However, challenges remain in handling heterogeneous data distributions, reducing communication overhead, and ensuring semantic consistency across data sources. Future Research Directions:
1) Cross-source feature alignment: Developing robust semantic mapping techniques will ensure that data from different origins can be processed consistently without compromising privacy.
2) Reducing communication costs: Optimizing distributed computing strategies will enhance the efficiency of federated learning by minimizing data exchange while maintaining performance.

\noindent\textbf{Computational Efficiency and Scalability: Optimizing Large-Scale Feature Engineering}
As datasets become increasingly complex, the computational demands of feature engineering grow, making many existing approaches impractical for real-time applications. Iterative feature generation and optimization further exacerbate these challenges, leading to significant computational bottlenecks. Future Research Directions:
1) Lightweight feature engineering algorithms: Developing approximation-based or low-rank decomposition methods can reduce computational burdens without sacrificing effectiveness.
2) Leveraging parallel computing and hardware acceleration: Distributed processing frameworks and GPU/TPU-based optimizations can improve the scalability of feature engineering, enabling real-time deployment.

\noindent\textbf{Large Language Models and Multimodal Feature Engineering: Expanding the Scope of Representation Learning}
The rapid advancement of LLMs and multimodal learning offers new possibilities for feature engineering by integrating structured tabular data with text, images, and other modalities. However, encoding tabular data effectively within LLM architectures remains a challenge, as does preserving the distinct characteristics of each modality during integration. Future Research Directions:
1) Developing specialized tabular embeddings: Designing encoding techniques tailored to tabular data will improve its compatibility with LLM-based frameworks.
2) Cross-modal feature alignment: Ensuring that multimodal data representations preserve semantic relationships is essential for achieving meaningful feature integration.
3) Efficient fine-tuning and knowledge transfer: Optimizing model adaptation techniques will enhance the practicality of LLM-based feature generation for tabular tasks.

\noindent\textbf{Future Outlook: Balancing Performance and Interpretability}
As feature engineering techniques become increasingly sophisticated, striking a balance between performance and interpretability is a pressing challenge. Complex transformation pipelines can obscure relationships between raw inputs and model outputs, making it difficult to ensure transparency, particularly in critical domains where accountability is essential. Future Research Directions:
1) Developing traceable feature generation frameworks: Logging and visualization tools can provide researchers with clear insights into how feature transformations affect model behavior.
2) User-friendly interpretability tools: Designing intuitive interfaces will enable non-technical stakeholders to understand feature engineering processes, facilitating broader AI adoption in real-world applications.

\vspace{-0.2cm}
\section{Conclusion}
Tabular data-centric AI is evolving with RL-based optimization and generative modeling playing a key role in feature engineering. This study examines the challenges and future directions in this domain, focusing on automation, interpretability, and scalability. By addressing these issues, we highlight both the limitations of existing methods and the transformative potential of AI-driven approaches.
Recent advances, particularly RL frameworks for adaptive feature selection and deep generative models for data augmentation, are redefining tabular data processing. These methods enhance autonomy and efficiency, bridging the gap between raw data and high-performance models. The integration of large language models and multimodal learning further expands the capabilities of tabular AI, enabling it to handle more diverse data sources.
Future research should refine RL-based optimization for stable training and efficient rewards while advancing generative techniques for high-quality, domain-adaptive features. As these challenges are addressed, tabular AI will continue to push automated machine learning forward, improving scalability and interpretability. This study aims to inspire further innovation, equipping researchers with new tools to maximize AI-driven feature engineering.
\newpage
\bibliographystyle{named}
\bibliography{survey,wdj}

\end{document}